\documentclass[10pt, conference, compsocconf,letterpaper]{IEEEtran}
\usepackage{color}
\usepackage{url}
\usepackage{hyperref}
\usepackage{algorithm,algcompatible}
\usepackage{algpseudocode}
\usepackage{amsmath}
\usepackage{float}

\newcommand{\runin}[1]{\vspace{.2em}\noindent\textbf{#1}}

   \usepackage[pdftex]{graphicx}
   \graphicspath{{./fig/}}
   \DeclareGraphicsExtensions{.pdf,.jpeg,.png}
\begin{document}
\title{Explaining Online Reinforcement Learning Decisions of Self-Adaptive Systems}


\author{\IEEEauthorblockN{Felix Feit, Andreas Metzger, Klaus Pohl}
\IEEEauthorblockA{paluno (The Ruhr Institute for Software Technology)\\
University of Duisburg-Essen; 
Essen, Germany\\
f.m.feit@gmail.com, andreas.metzger@paluno.uni-due.de, klaus.pohl@paluno.uni-due.de}
}

\maketitle

\begin{abstract}
Design time uncertainty poses an important challenge when developing a self-adaptive system. 
As an example, defining how the system should adapt when facing a new environment state, requires understanding the precise effect of an adaptation, which may not be known at design time. 
Online reinforcement learning, i.e., employing reinforcement learning (RL) at runtime, is an emerging approach to realizing self-adaptive systems in the presence of design time uncertainty. 
By using Online RL, the self-adaptive system can learn from actual operational data and leverage feedback only available at runtime. 
Recently, Deep RL is gaining interest.
Deep RL represents learned knowledge as a neural network whereby it can generalize over unseen inputs, as well as handle continuous environment states and adaptation actions. A fundamental problem of Deep RL is that learned knowledge is not explicitly represented. 
For a human, it is practically impossible to relate the parametrization of the neural network to concrete RL decisions and thus Deep RL essentially appears as a black box. 
Yet, understanding the decisions made by Deep RL is key to (1) increasing trust, and (2) facilitating debugging. 
Such debugging is especially relevant for self-adaptive systems, because  the reward function, which quantifies the feedback to the RL algorithm, must be defined by developers.
The reward function must be explicitly defined by developers, thus introducing a potential for human error. 
To explain Deep RL for self-adaptive systems, we enhance and combine two existing explainable RL techniques from the machine learning literature. 
The combined technique, XRL-DINE, overcomes the respective limitations of the individual techniques.  
We present a proof-of-concept implementation of XRL-DINE, as well as qualitative and quantitative results of applying XRL-DINE to a self-adaptive system exemplar.
\end{abstract}

% \fi
\IEEEpeerreviewmaketitle

\section{Introduction}
\label{sec:Introduction}

\runin{A self-adaptive system} can modify its own structure and behavior at runtime based on its perception of the environment, of itself and of its requirements~\cite{Weyns2020}.
One key element of a self-adaptive system is its \emph{self-adapt\-ation logic} that encodes when and how the system should adapt itself.
When developing the adaptation logic, developers face the challenge of \emph{design time uncertainty}~\cite{CalinescuMPW20,MetzgerEtAl2022,WeynsEtAl22}.
To define \emph{when} the system should adapt, they have to anticipate all potential environment states.
However, this is infeasible in most cases due to incomplete information at design time. 
As an example, the concrete services that may be dynamically bound during the execution of a service orchestration and thus their quality characteristics are typically not known at design time.
To define \emph{how} the system should adapt itself, developers need to know the precise effect an adaptation action has.
However, the precise effect may not be known at design time.
As an example, while developers may know in principle that enabling more features will negatively influence the performance, exactly determining the performance impact is more challenging.
A recent industrial survey identified optimal design and design complexity together with design time uncertainty to be the most frequently observed difficulties in designing self-adaptation in practice~\cite{WeynsEtAl22}.

\runin{Online reinforcement learning (Online RL)} is an emerging approach to realize self-adaptive systems in the presence of design time uncertainty.
Online RL means that reinforcement learning~\cite{sutton_reinforcement_2018} is employed at runtime (see~\cite{MetzgerEtAl2022} for a discussion of existing solutions).
The self-adaptive system thereby can learn from actual operational data and thus leverages information only available at runtime.
A recent survey indicates that since 2019 the use of learning dominates over the use of predetermined and static policies or rules~\cite{PorterFD20}.

Online RL aims at learning suitable adaptation actions via the self-adaptive system's interactions with its initially unknown environment~\cite{CAISE2020}. 
During system operation, the RL algorithm receives a numerical reward based on actual runtime monitoring data for executing an adaptation action.
The reward expresses how suitable that adaptation action was in the short term. 
The goal of Online RL is to maximize the cumulative reward. 

Initially, research on self-adaptive systems leveraged RL algorithms that represent learned knowledge as a so-called \textit{value function}~\cite{CAISE2020}.
The value function quantifies how much cumulative reward can be expected if a particular adaptation is chosen in a given environment state. 
Typically, this value function was represented as a table. 
However, such tabular approaches exhibit key limitations. 
First, they require a finite set of environment states and a finite set of adaptations and thus cannot be directly applied to continuous state spaces or continuous action spaces. 
Second, they do not generalize over neighboring states, which leads to slow learning in the presence of continuous environment states~\cite{CAISE2020}.

\textit{Deep reinforcement learning} (\emph{Deep RL}) addresses these disadvantages by representing the learned knowledge as a neural network. 
Since neural network inputs are not limited to elements of finite or discrete sets, and neural networks can generalize well over inputs, deep RL has shown remarkable success in different application areas. 
Recently, Deep RL is also being applied to self-adaptive systems~\cite{moustafa2018deep,wang2019adaptive,CAISE2020}.
%% ADD
%Large-scale and adaptive service composition using deep reinforcement learning

\runin{A principal problem of Deep RL} is that learned knowledge is not explicitly represented.
Instead, it is ``hidden'' in the parametrization of the neural network. 
For a human, it is practically impossible to relate this parametrization to concrete RL decisions.
Deep RL thus essentially appears as a black box~\cite{puiutta_explainable_2020}.
Yet, understanding the decisions made by Deep RL systems is key to (1) increase trust in these systems, and (2) facilitate their debugging~\cite{miller_explanation_2019,Dusparic_2022}.

Facilitating the debugging of Deep RL is especially relevant for self-adaptive systems, because Online RL does not completely eliminate manual development effort.
Since developers need to explicitly define the reward function, this introduces a potential source for human error.
%% too much for intro, maybe later on?
%// naturally reflects the xx in the environment (da war doch mal ein Zitat)
%Often, reward functions are derived from a utility function that balances the various, often conflicting, goal dimensions.
%But, getting such a utility function "right" in a sense that it accurately reflects the trade-off among different goal dimensions is a challenge.
%Further, recent findings show that modeling the reward function too closely to reality may slow down learning ((BPM Paper!!!)).

To explain Deep RL systems, various \textit{Explainable Reinforcement Learning} (\emph{XRL}) techniques were recently put forward in machine learning research~\cite{puiutta_explainable_2020, heuillet_explainability_2020}. 
%However, with the exception of~\cite{UllauriGBZZBOY22}, XRL approaches have not yet been applied and studied for self-adaptive systems.
Here, we set out to answer the question how existing XRL techniques can be applied for the explainability of Online RL for self-adaptive systems.
We follow XRL literature and use "explainable" to also include "interpretable", even though one may consider "interpretable" only as basis for "explainable".

\runin{Our contribution }is to enhance and combine two existing XRL techniques from the literature: \textit{Reward Decomposition}~\cite{juozapaitis_explainable_2019} and \textit{Interestingness Elements}~\cite{sequeira_interestingness_2020}.
Reward Decomposition uses a suitable decomposition of the reward function into sub-functions to explain the short-term goal orientation of RL, thereby providing contrastive explanations.

Reward composition is especially helpful for the typical problem of adapting a system while taking into account multiple quality goals.
Each of these quality goals could then be expressed as a reward sub-function.
%These explanations help to understand which goal a chosen adaptation action contributes to.
However, no indication for the explanation's relevance is provided, but instead it requires manually selecting relevant RL decisions to be explained. 
In particular when RL decisions are taken at runtime, which is the case for Online RL for self-adaptive systems, monitoring all explanations to identify relevant ones introduces cognitive overhead for developers.
In contrast, Interestingness Elements collect and evaluate metrics at runtime to identify relevant moments of interaction between the system and its environment. 
%Interestingness Elements thereby facilitate selecting relevant actions.
However, for an identified relevant moment of interaction, it does not provide explanations whether the system's decision making behaves as expected and due to the right reasons.

Our technique,\emph{ XRL-DINE}, combines the two aforementioned techniques to overcome their respective limitations. 
XRL-DINE provides detailed explanations at relevant points in time by computing and visualizing so called \emph{Decomposed INterestingness Elements} (\emph{DINEs}).
We introduce three types of DINEs: ``Important Interaction'', ``Reward Channel Extremum'', and ``Reward Channel Dominance''.

We prototypically implement and apply XRL-DINE to the self-adaptive system exemplar SWIM -- a self-adaptive web application~\cite{moreno_swim_2018} -- to serve as proof a concept and to provide qualitative and quantitative results.

Sect.~\ref{sec:Foundations} provides foundations as basis for introducing XRL-DINE in Sect.~\ref{sec:Approach}.
Sect.~\ref{sec:Evaluation} describes the proof-of-concept implementation of XRL-DINE as well as its qualitative and quantitative evaluation, while Sect.~\ref{sec:discussion} discusses limitations.
Sect.~\ref{sec:RelatedWork} relates XRL-DINE to existing work.
\section{Foundations}
\label{sec:Foundations}

\subsection{Online RL for Self-adaptive Systems}

\begin{figure}[hbtp]
\centering
\includegraphics[width=.30\textwidth]{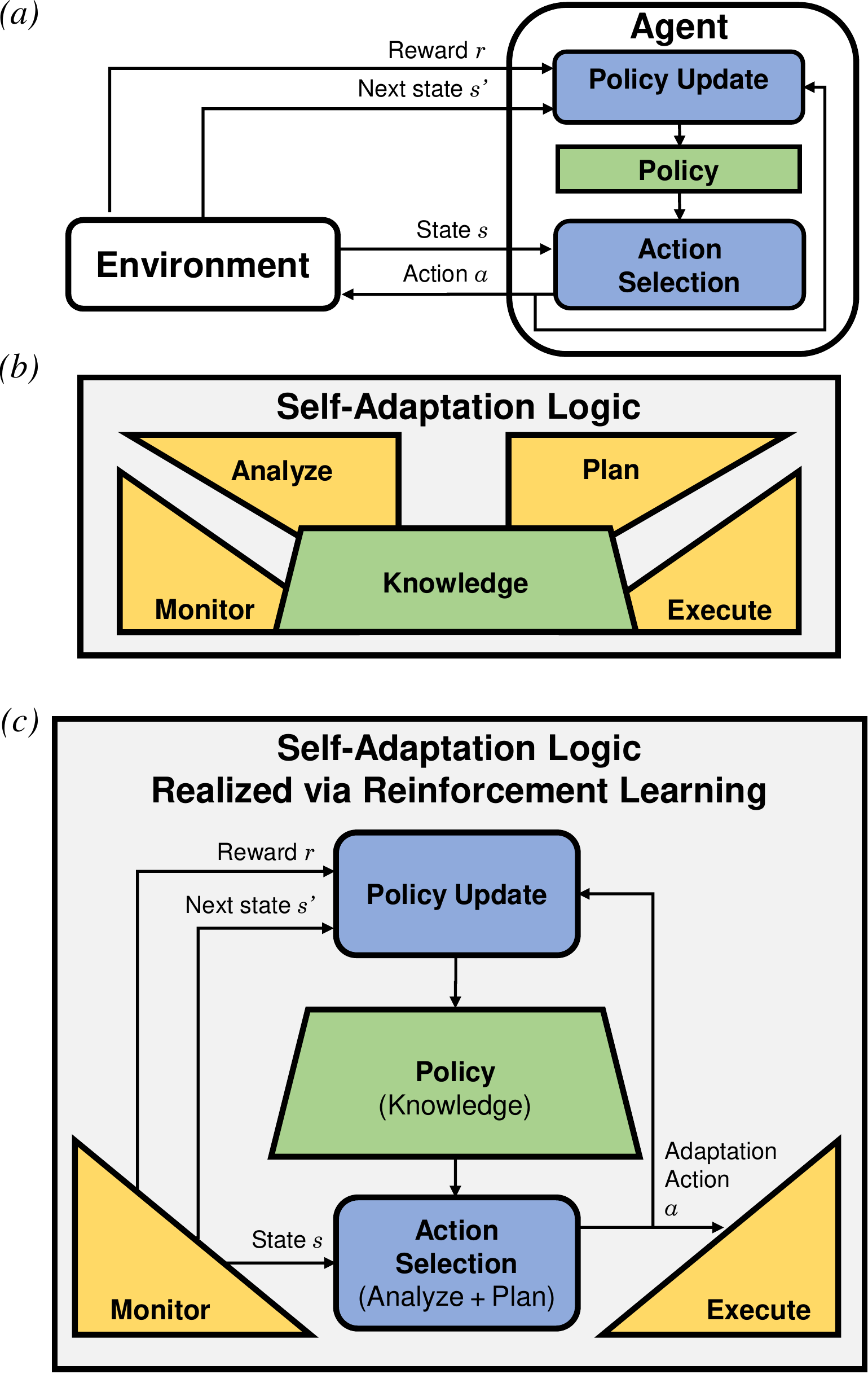}
\vspace{-.5em}
\caption{RL, MAPE-K, and their integration (adapted from~\cite{MetzgerEtAl2022})}
\vspace{-.5em}
\label{fig:architecture}
\end{figure}

\runin{Reinforcement Learning (RL)} aims to learn an optimal action selection policy via a system's (called agent in RL) interactions with its initially unknown environment~\cite{sutton_reinforcement_2018}.
As  shown in Fig.~\ref{fig:architecture}\emph{(a)}, the 
agent finds itself in environment state $s$ at a given time step.
The agent then selects an action $a$ (from its set of potential adaptation actions) and executes it. 
As a result, the environment transitions to the next state $s'$ 
and the agent receives a reward $r$ for executing the action.
The reward $r$ together with the information about the next state $s'$ are used to update the action selection policy of the agent.
The goal of RL is to maximize the cumulative reward.
\emph{Online RL} applies RL during system operation, where actions have an effect on the live system, resulting in reward signals based on actual monitoring data~\cite{CAISE2020}.

This paper focuses on explaining value-based Deep RL approaches for self-adaptive systems.
The reason is that the employed XRL technique of Reward Decomposition requires using value-based RL.
In value-based Deep RL, the policy depends on a learned action-value function $Q(S,A)$, which gives the expected cumulative reward when executing adaptation action $A$ in state $S$.
Value-based Deep RL uses a neural network to approximate $Q(S,A)$~\cite{sutton_reinforcement_2018}. 

\runin{A self-adaptive system }can conceptually be structured into two main elements~\cite{Weyns2020}: the \emph{system logic} and the \emph{self-adaptation logic}.
To understand how RL can be leveraged for realizing the self-adaptation logic, we use the well-established MAPE-K reference model for self-adaptive systems~\cite{Weyns2020}.
As depicted in Figure~\ref{fig:architecture}\emph{(b)}, MAPE-K structures the self-adaptation logic into four main conceptual activities that rely on a common \emph{knowledge} base.
These activities \emph{monitor} the system and its environment, \emph{analyze} monitored data to determine adaptation needs, \emph{plan} adaptation, and \emph{execute} these adaptations at runtime.

\runin{Online-RL for self-adaptive systems} integrates the elements of RL into the MAPE-K loop as shown in Fig.~\ref{fig:architecture}\emph{(c)}.
 %in order to facilitate learning effective adaptations at runtime.
For a self-adaptive system, ``agent'' refers to the self-adaptation logic of the system and ``action'' refers to an adaptation action~\cite{MetzgerEtAl2022}.
In the integrated model, \emph{action selection} of RL takes the place of the \emph{analyze} and \emph{plan} activities of MAPE-K.
The learned \emph{policy} takes the place of the self-adaptive system's \emph{knowledge} base. 
At runtime, the policy is used by the self-adaptation logic to select  an adaptation action $a$ based on the current state $s$ determined by \emph{monitoring}.
%Action selection determines whether there is a need for an adaptation (given the current state) and plans (i.e., selects) the respective adaptation action to \emph{execute}.
Using the policy, either a specific adaptation actions is selected and then \emph{executed}, or no adaptation is executed, which means the system is left in its current state. 

\subsection{Explainable RL}

Below we summarize the two XRL techniques that serve as basis for our combined approach XRL-DINE.

\runin{Reward Decomposition}, originally proposed to improve learning performance, was exploited by Juozapaitis et al. for the sake of explainability~\cite{juozapaitis_explainable_2019}.
Reward Decomposition divides the reward function of the RL agent into several sub-functions, called \emph{reward channels}, which reflect a different aspect of the learning goal.
For each of the sub-functions a separate RL agent, we call \emph{sub-agent}, is trained.
To select a concrete action,  an aggregated value-function is computed by accumulating the values for each of the actions proposed by the different channels. 
The resulting aggregated value-function is then used for action selection, while trade-offs in decision making made by the composed agent become observable via the reward channels. 

An example is a self-adaptive web application that may add more cloud instances to respond to high user load. 
The additional costs incurred by the additional cloud instance have to be weighed against reduced user satisfaction due to slower response times. 
By using two reward channels, one for a positive reward due to increased user satisfaction and one for a negative reward (i.e., penalty) due to infrastructure costs, this trade-off can be made explicit and thus observable.

\runin{Interestingness Elements} were proposed by Sequiera et al.~\cite{sequeira_interestingness_2020}.
By analyzing state transition data collected at  runtime, infrequent situations and thereby relevant points of agent-environment interactions are determined. 
%By tracking state transitions, non-frequent situations are identified.

In our self-adaptive web application example from above, an Interestingness Element may be determined in an unfamiliar situation for the RL agent due to a non-anticipated, abrupt increase in user requests (i.e., workload). 

%%% NEEDED?
%Furthermore, the authors found out in a user study that good and bad interaction moments of the agent with the environment have to be presented so that the user does not get a distorted picture of its capabilities and limitations~\cite{sequeira_interestingness_2020}.

%%%% do we need???
%Each metric can also be collected externally if the algorithm does not provide it already: $n(s)$, $n(s,a)$, and $n(s,a,s')$ can be collected by employing a set of counters that keep track of the according state transitions. $\hat{P}(s'|s,a)$ can then be derived from these counters by calculating $\hat{\mathrm{P}}\left(s^{\prime} \mid s, a\right)=n\left(s, a, s^{\prime}\right) / n(s, a)$. Finally, $Q$ and $V$ can be retrieved by running an on-policy value-based reinforcement learning algorithm (see Section~\ref{sec:reinforcement_learning}) in parallel to the actual algorithm.

\section{Approach}
\label{sec:Approach}

XRL-DINE combines the above two XRL  techniques to overcome their respective limitations.
Interestingness Elements are used to select relevant interactions, which are then explained contrastively by using Reward Decomposition. 
The resulting DINEs (Decomposed Interestingness Elements) show relevant alternative decisions of the RL agent. 
%If there is no relevant alternative decision at a given timestep, no DINE is displayed. 
Below we explain the three types of DINEs and show their visualization in the XRL-DINE dashboard.

\subsection{``Important Interaction'' DINE} 
%\todo{following § needs more explanation}
%The linking of \textit{Interestingness Elements} with \textit{Reward Decomposition} to form \textit{Decomposed Interestingness Elements} occurs at two different levels of abstraction. 
%First, for \textit{Important Interactions} the Interestingness Element \emph{certain actions} is solely used to extract relevant, contrastive actions that are then explained by generic, natural language Reward Decomposition explanations. 
%Second, \textit{Reward Channel Extremes} are computed separately for each reward channel and are thus more tightly coupled.
This DINE is a modification of the ``certain/uncertain action'' Interestingness Element proposed in~\cite{sequeira_interestingness_2020}. 
The authors measure whether the agent in a given state chooses a wide range of different actions, thus being considered ``uncertain'', or whether the agent almost always chooses the same action, thus being considered ``certain''. 

However, applying this original approach to value-based Deep RL, which we focus on in this paper (see Sect.~\ref{sec:Foundations}), provides limited usefulness.
%The original approach only provided useful insights, as long as the RL algorithm had not converged. 
After convergence of the learning process, always the same explanations would be given, as always the same action is chosen for a given state (due to the greedy selection from the $Q$ values).

To facilitate RL explanations after convergence, we modify the calculation of the Interestingness Element.
Instead of relying on the inequality of action selection, we compute the inequality of the normalized action-values. 
This computation follows the same formula proposed in~\cite{sequeira_interestingness_2020} for computing the inequality of a probability distribution. 
However, since this modification  slightly changes the semantics of the element, the term \textit{Important Interaction} is adopted in this work instead of \emph{certain} action. 
The adjective \textit{important} expresses that something has \emph{great effect or influence}, which follows the intuition behind the derivation of Important Interactions: A sub-agent considers an action \textit{important} if its action-value is very different relative to the action-values of all alternative actions. 
Thereby, the actual absolute action-value is not of great importance, since the agent may be in a disproportionately good (or bad) part of the state-action space, where there is a naturally high (or low) expected cumulative reward. 
Calculating the inequality of action-values for a given state captures the internal relative weighting of the agent.

To link this modified Interestingness Element to Reward Decomposition, the inequality of action-values is calculated for each reward channel. 
Only if inequality is found for at least one of these reward channels and the action of the aggregated agent does not correspond to the action that the sub-agent would have chosen, the action is identified as an Important Interaction. 
This combination yields contrastive explanations in which the sub-agent’s desired action represents the contrasting action. 

To tune the number of DINEs generated and thus cope with potential cognitive overload, XRL-DINE offers setting a threshold parameter $\rho$.
This parameter controls the necessary level of inequality at which a decision is labeled as \textit{important} to a sub-agent. 
The lower this threshold is set, the more elements will be generated with varying degrees of importance. 
In contrast, for high thresholds, none or only a few, but more relevant elements are identified.

To facilitate scalability for a higher number of actions and reward channels, one may generate \textit{Minimal Sufficient Explanations}~\cite{juozapaitis_explainable_2019}, which are the minimal sets of sub-agents that are necessary for the aggregated action to be chosen. 
They would be generated by ranking the action values of the sub-agents and computing the outcomes by increasingly adding the action values of the sub-agents until they outweigh the action value of the best alternative action.

\subsection{``Reward Channel Extremum'' DINE}
This DINE is based on the "Minima / Maxima Situations" of Interestingness Elements indicating local extrema of the state-value function $V(S)$, which gives the expected reward in state $s \in S$. 
These extrema represent actions directly followed only by states with a higher state-value (local minimum) or lower state-value (local maximum)~\cite{sequeira_interestingness_2020}. 
Such local extrema  help identifying RL's reasoning in potentially critical states~\cite{Dusparic_2022}.
``Reward Channel Extremum'' DINES can be linked to reward decomposition in two ways: 
First, a generic reward decomposition explanation can be displayed when the overall agent reaches a local maximum or minimum. 
Second, developers are informed when one of the sub-agents is in a local maximum or minimum. 

"Reward Channel Extremum" DINEs provide developers with an overview of how (1) the overall system and (2) each individual sub-agent evaluates a sequence of decisions and what actions are taken by the agent to leave a local reward minimum as quickly as possible or to maintain a high cumulative reward after a local reward maximum.

Even though the intuition behind ``Reward Channel Extremum'' DINEs is straightforward, deriving these DINEs requires predicting the next state $s'$ for each possible action $a \in \mathcal{A}$ in state $s$, in order to be able to determine maxima in an online fashion.
The value-based deep RL approaches we focus in this paper (see Sect.~\ref{sec:Foundations}) belong to the class of model-free RL algorithms.
This means that they do not use a model of the environment.
Such an environment model must therefore be approximated by XRL-DINE.
Such an approximation depends on the concretely chosen RL algorithm and is thus discussed in Sect.~\ref{sec:Prototype}. 

To generate the ``Reward Channel Extremum'' DINEs, at each time step for all possible actions $a \in \mathcal{A}$, and for the current state $s$ the next states are predicted using the approximated environment model. 
For these $|\mathcal{A}|$ predicted next states, all sub-agents then compute the state-value respectively. 
Since the sub-agents only approximate the action-value function, the state-value $V(s)$ is derived from the action-value function by choosing the action-value of the greedy action (i.e., the action with the highest $Q$ value):
\vspace{-.4em}
$$
V(s)=\max _{a \in \mathcal{A}} {Q}(s, a)
\vspace{-.2em}
$$

If the state-values of all predicted next states are worse than the state-values of the current state, a local reward maximum is reached. 
If all predicted next states are better, a local reward minimum is reached. 
Since such local extrema may occur very often, especially for a small number of discrete actions, we propose using a threshold $\phi$.
To determine a local minimum, the best action-value of the current state must be $\phi$ lower than the lowest action-value of the following state, and vice versa. 
The frequency and amount of information shown to the developers can thereby be controlled by $\phi$.

\subsection{``Reward Channel Dominance'' DINE} 
This DINE originates in Reward Decomposition research.
The DINE gives compact information of the influence that each sub-agent has on each possible action. 
Here, we include it among the DINEs because it enhances the information provided by the other two DINEs from above. 

We propose two types of ``Reward Channel Dominance'' DINEs: 
\emph{Absolute Reward Channel Dominance }gives the action-values of each sub-agent for a given state. 
Since the action-values are not bound, the values can also be negative and also vary widely between sub-agents. 
To better explain the actual contribution of each sub-agent to the aggregated decision, we propose converting the absolute channel dominance values into relative channel dominance values. 
%This \textit{Relative Reward Channel Dominance} makes the agent's internal deliberation process explicit, helping the user (or supervisor) to understand the decisions made. 
To compute this, for each sub-agent, the value of the worst action is subtracted from the action-values of all other actions. 
This results in (1) all values being positive, and (2) limiting the imbalance of contributions only to the \emph{decisive} portion. 
However, since the conversion of absolute to relative dominance values may involve loss of information, both types of elements are provided to the developers.

\subsection{Visualization in XRL-DINE}
We display the outcomes of XRL-DINE using a visual dashboard. 
It allows navigating the decision trajectory and thereby also facilitates investigating explanations of past interactions. 
The purpose of this type of visualization is to preserve the respective advantages of the two combined explanation techniques:
Being able to gain an understanding of the RL agent's higher-level behavior, while also being able to investigate specific actions of interest.

\begin{figure}[h]
    \centering
    \includegraphics[width=\columnwidth]{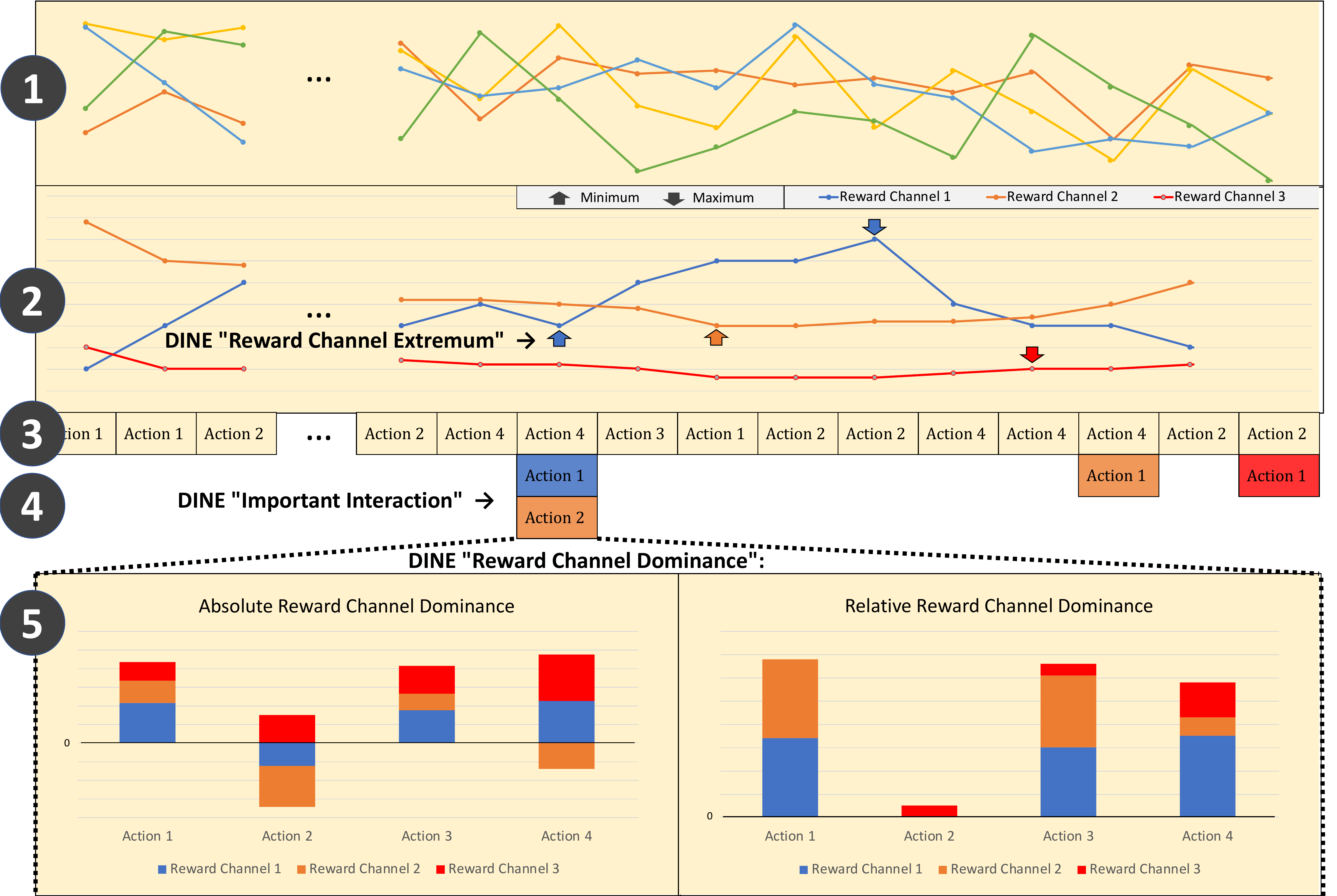}
	\vspace{-1.5em}
    \caption{XRL-DINE Dashboard}
	\vspace{-1.5em}
    \label{fig:proposed_dashboard_sketch}
\end{figure}

The principal layout of the dashboard is shown in Fig.~\ref{fig:proposed_dashboard_sketch}.
It  follows an interaction concept that allows visual data exploration~\cite{keim_visual_2001} and is centered around time progression as main axis. 
The different visual elements are linked, so that hovering over one element highlights the information of other elements for the same time step. 
%By selecting a specific time step, the reward dominance values calculated for that time step and are displayed as barcharts. 
We took into account previous experience on user-centered design~\cite{endsley2003designing}.
To address cognitive overload, we (1) already reduce this by computing DINEs, and (2) only show relevant information when needed.
To avoid the requisite memory trap (i.e., relying on limited short-term memory), we show the complete historic information in a compact representation.

The dashboard shows the following: 
\textbf{(1)~State progression}, which is represented using a line graph. 
Each line in the dashboard represents a Z-score standardized state variable. 
Standardization is necessary to display differently scaled variables. 
\textbf{(2)~Received rewards progression} for each reward channel together with the ``Reward Channel Extremum'' DINEs.
\textbf{(3)~Trajectory of selected actions}, where adaptation actions chosen by the composed RL agent are shown.
\textbf{(4)~Important Interactions}, which shows the ``Important Interaction'' DINEs. 
The background color corresponds to the color of the particular reward channel for which the contrastive action is considered important. 
%If multiple reward channels suggest an important action different from the selected action, they are stacked on top of each other.
The dashboard also provides textual explanations for these important interactions (not shown in the figure).
\textbf{(5)~Reward Channel Dominance}, which is displayed in a stacked column chart. 
Each column represents a possible adaptation action. 
The reward channels are shown in the same colors as for the other DINEs.

\section{Validation}
\label{sec:Evaluation}

Following established practices in self-adaptive systems research~\cite{PorterFD20}, we perform real-world experiments to validate XRL-DINE.
This means we implement a system and then subject it to approximately realistic conditions.

\subsection{Proof-of-concept Implementation of XRL-DINE}
\label{sec:Prototype}

\begin{figure*}[t]
    \centering
    \includegraphics[width=1.6\columnwidth]{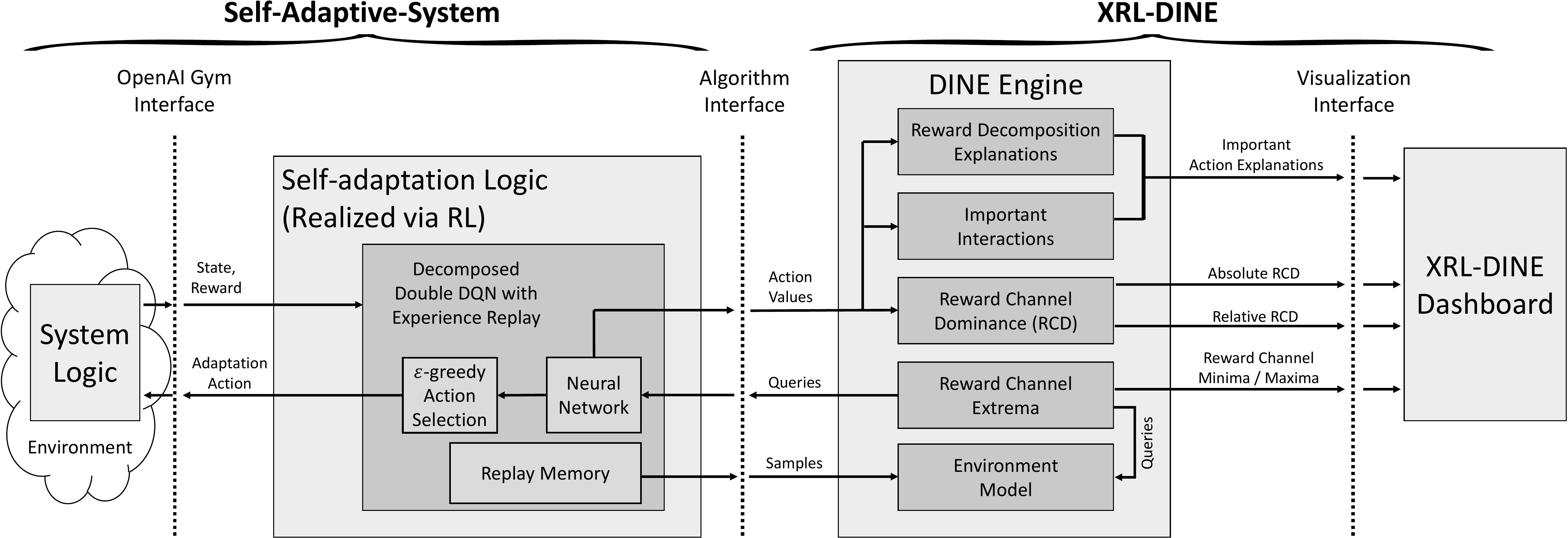}
    \caption{Architecture of Prototypical Implementation of XRL-DINE}
    \vspace{-1em}
    \label{fig:interfaces}
\end{figure*}

To provide a proof of concept and to demonstrate the effectiveness of XRL-DINE, we prototypically implemented it for a concrete Deep RL algorithm\footnote{Code is available via \url{https://git.uni-due.de/rl4sas/xrl-dine}}.
In our prototype, we use \textit{Double Deep Q-Networks with Experience Replay}
as a state-of-the-art representative of value-based algorithms~\cite{hasselt_deep_2016}. 
Fig.~\ref{fig:interfaces} shows the main components of the XRL-DINE implementation (the XRL-DINE engine and dashboard) and how they connect with the implementation of the self-adaptive system using Deep RL.

In the used Deep RL algorithm, the data observed during agent-environment interactions are not used directly for training, but are first written to a replay memory. 
At each time step, a predefined number of samples (called \textit{batch}) is randomly taken from the replay memory and the current bootstrapping target is calculated for each sample.
The bootstrapping target is an approximation of the expected cumulative reward of an action. 
It is formed as the weighted sum of the immediate reward received for the chosen action and the action-value of the best possible action in the successor state.
For the calculation of the bootstrapping target the other sub-agents are considered as well. 
We thus extend the Deep RL algorithm to the decomposed case with the update rule proposed in~\cite{juozapaitis_explainable_2019}. 
%Using the target and the predicted action-values we form the loss and then perform backpropagation to adjust the neural network weights to (gradually) minimize the loss.

As shown Fig.~\ref{fig:interfaces}, the Deep RL algorithm that realizes the self-adaptation logic is linked to the XRL-DINE Engine as follows.
First, the action values are transmitted to the XRL-DINE engine at each time step via a callback. 

Second, as explained in Sec.~\ref{sec:Approach}, an approximated environment model is needed for computing the ``Reward Channel Extremum'' DINEs.
In our implementation, we create such a model using supervised learning by exploiting the contents of the replay memory as a labeled dataset.
The replay memory contains past transitions in the form $(s, a, r, s')$ (i.e., State-Action-Reward-Next State~\cite{sutton_reinforcement_2018}). 
Using $s$ and $a$ as input and $s'$ as output, we train a generic feed-forward neural network as a universal nonlinear function approximator. 
During the initialization of the XRL-DINE engine, links to the replay memory (for training the environment model) and the neural networks of the individual RL sub-agents (for evaluating successor states) are passed by reference.

This implementation of the self-adaptation logic is connected to the self-adaptive system and its environment using the OpenAI Gym interface, a widely used interface between RL agents and their environments\footnote{For details see \url{https://www.gymlibrary.ml/content/api/}}.
Classes that implement this interface offer a ``step'' method, to which the action to be executed is passed. 
This action is then executed and the next state and the reward received are returned. 
Since Reward Decomposition requires a vector of reward values instead of the default scalar reward value provided by the OpenAI Gym interface, we have slightly modified the interface.

\subsection{Application to Self-adaptive System Exemplar}
\label{sec:swim}

We apply XRL-DINE to the SWIM exemplar, which is one of the self-adaptive system exemplars provided by the SEAMS community~\cite{moreno_swim_2018}. 
SWIM simulates a self-adaptive multi-tier web application.
It closely replicates the real-time behavior of an actual web application, while allowing to speed up the simulation to cover longer periods of real time.
%The SWIM exemplar is an instance of the auto-scaling problem in which the objective is to provision resources to satisfy conflicting business goals. Specifically, self-adaptive logic must be implemented that maximizes a given utility function, despite varying system load.

SWIM has different monitoring metrics to determine the state of the system.
These metrics include the request arrival rate (i.e., ``workload'') as well as the average throughput and response time.
As all three environment variables are continuous, SWIM's state space is continuous.
Therefore, tabular RL solutions cannot directly be applied to the exemplar.

SWIM can be adapted as follows:
(1)~additional web servers can be added / removed, resulting in the load being distributed across more / fewer servers; (2)~the proportion of requests for which optional, computationally intensive content is generated (e.g., via recommendation engines) can be modified by setting a so called dimmer value. 
While adaptations of type (1) have an impact on costs, adaptations of type (2) have an  impact on revenue.

To apply XRL-DINE, we define the following decomposed reward function, consisting of three reward channels:
\vspace{-1.5em}
\begin{align*}
R_{\mathrm{total }}=
a \cdot R_{\mathrm{user\_satisf.}} + 
b \cdot R_{\mathrm{revenue }} + 
c \cdot R_{\mathrm{costs }}
\end{align*}
\vspace{-1.5em}

The weights were selected experimentally according to two criteria. 
First, no reward channel should dominate the other two reward channels to such an extent that the decisions of the other sub-agents have no influence on the choice of actions. 
This criterion reflects the basic assumption of DINEs that multiple sub-agents should find the best possible trade-off. 
The second, subordinate criterion is to conform as closely as possible to the original utility function. 
Specifically, the parameters were chosen such that \textit{User Satisfaction} has the highest influence ($a = 4$), \textit{Revenue} has the second highest influence ($b = 2$), and \textit{Costs} has the lowest influence ($c = 1$). 

The three reward sub-functions are defined as follows:
\vspace{-.1em}
\begin{align*}
R_{\text {user\_satisf.}}= \begin{cases}0.5 & : x \leq 0.02 \\ -0.5-\frac{x-1}{20} & : x \geq 1 \\ 0.5- \frac{x-0.02}{0.98} & : \text { otherwise }\end{cases}
\vspace{-1em}
\end{align*}
%\begin{align*}
%R_\textrm{user\_satisf.} = 0.5 &\textrm{ if } x \leq 0.02; \\
%-0.5-(x-1)/20 &\textrm{ if }  x \geq 1; \\
%0.5- (x-0.02)/0.98 &\textrm{ otherwise} 
%\vspace{-1em}
%\end{align*}
In $R_\textrm{user\_satisf.}$,  the perceived user satisfaction depends on the average latency $x$.
This utility function is provided as part of the SWIM exemplar.
\vspace{-.2em}
\begin{align*}
R_\textrm{revenue}= \tau \cdot a \cdot\left(d \cdot R_{O}+(1-d) \cdot R_{M}\right)
\vspace{-.5em}
\end{align*}
In $R_\textrm{revenue}$, $\tau$ is the length of the time interval between two consecutive time steps, $a$ is the average arrival rate of requests and $d$ is the current dimmer value. 
$R_{M}$ is the reward obtained by processing a request without optional content. 
$R_{O}$ is the reward obtained by processing a request with optional content. 
%The term $d \cdot R_{O}+(1-d) \cdot R_{M}$ thus represents the average reward controlled by the dimmer for each request.
\vspace{-1.5em}
\begin{align*}
R_\textrm{costs} = -(\tau \cdot c \cdot s)
\end{align*}
\vspace{-1.5em}

In $R_\textrm{costs}$, $s$ indicates the number of servers currently in use. 
The parameter $c$ models the cost of using one server.
This means, $R_\textrm{costs}$ is higher the fewer servers are used.

Finally, to prevent illegal actions, we explicitly punish them by adding a fixed penalty to each reward channel. 
In addition, all actions that cause a change in the system state (i.e., all actions except ``No Adaptation'') are penalized to account for increased computational overhead and to guide the aggregated agent towards a less jerky policy.

\begin{figure*}[ht]
    \centering
    	$
		\begin{array}{cc}
    \includegraphics[width=.45\textwidth]{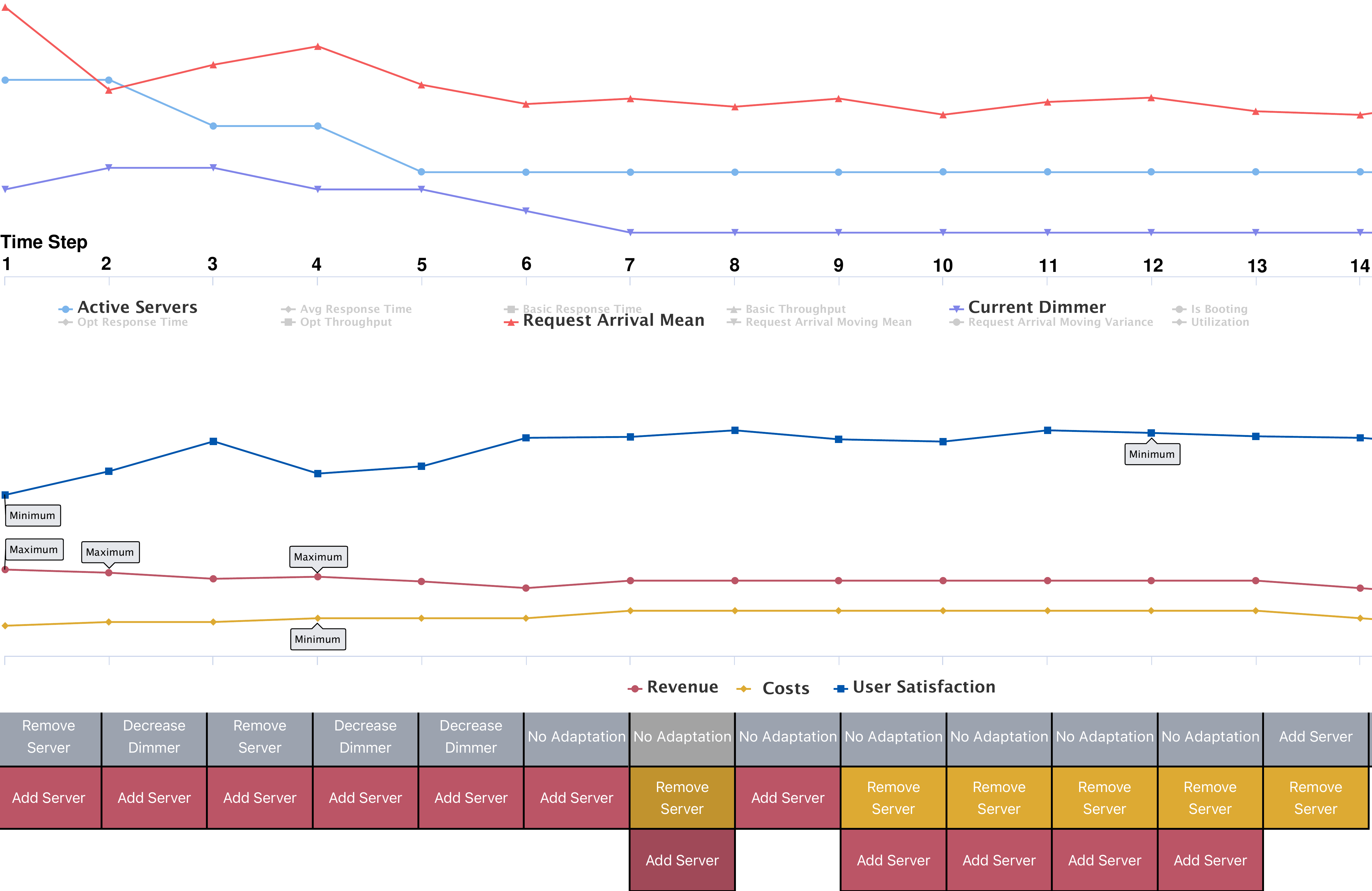} &
    	\includegraphics[width=.35\textwidth]{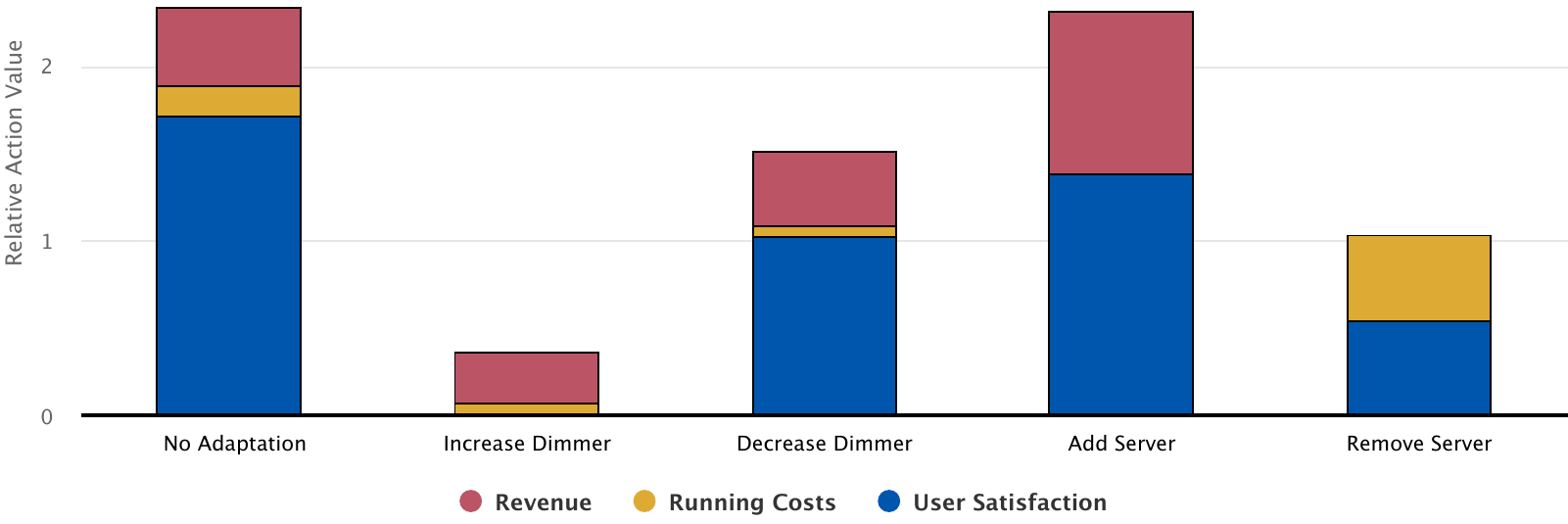} 		
\end{array}$
    \vspace{-.5em}
    \caption{XRL-DINE dashboard for SWIM exemplar}
   	\vspace{-1.5em}

    \label{fig:example_3_top}
\end{figure*}

%We apply the \textit{Decomposed Double Deep Q-Learning Agent} explained above to the SWIM environment. 

\subsection{Qualitative Results}
\label{sec:apply}

When applying XRL-DINE to the SWIM exemplar, we identified a set of representative example explanations that show the insights that can be gained by using XRL-DINE but also its limitations.
Due to space limitations, we report on a subset of examples showing the insights to be gained and briefly discuss limitations in Sect.~\ref{sec:discussion}.

We chose example explanations generated after initial convergence of the RL agent (and also including after training the environment model), which allows determining whether the self-adaptive system will behave as may be expected by domain experts.
The left hand side of Fig.~\ref{fig:example_3_top} shows the contents of the main part of the XRL-DINE dashboard for the example\footnote{Interactive dashboard available via \url{https://git.uni-due.de/rl4sas/xrl-dine}}.
This figure shows an excerpt of the overall learning process.
For convenience, we start numbering the time steps with 1, even though the time step on the leftmost side actually occurred at step 22,575.
The figure shows how the RL agent responds to an increase in user requests. 
This increase is reflected in the red curve in the XRL-DINE dashboard. 
This curve represents the normalized duration between incoming requests. 
The lower this value, the higher the request rate (i.e., the request rate is the inverse of the normalized duration between incoming requests). 
At time step 2, there is a short peak in the request rate, which then decreases again\footnote{For sake of readability, we do not use the actual time step numbers}. 
Between time steps 4 and 6, the request rate increases again and then more or less remains at this higher level for the remainder of the example. 
The number of active servers is 10 at time step 1, but is lowered by the RL agent down to 8 servers by time step 5. 
The dimmer value is at 0.1 at time step 1 and is reduced by the RL agent to 0 by time step 7.

The curve of the \textit{User Satisfaction} reward channel depicts an increase between time steps 1 and 3, which decreases again due to the shutdown of two servers after time step 3. 
By lowering the dimmer value in time steps 4 and 6 and due to a more or less stable request rate, the reward for the \textit{User Satisfaction} reward channel increases and stabilizes at a high level from time step 6 onward. 
Lowering the dimmer value also causes the reward for the \textit{Revenue} channel to decrease and stabilize at a lower level. 
In contrast, the reward for \textit{Running Costs} increases, since two less servers need to be operated. 
By choosing \emph{No Adaptation} between time steps 7 and 13, all reward channels receive a relative boost, as the $-0.1$ penalty for choosing any action except \emph{No Adaptation} is no longer received. In summary, the aggregated agent's strategy is to respond to an increase in the request rate by lowering the dimmer, thus trading a gain in \textit{User Satisfaction} and \textit{Running Costs} for a loss of \textit{Revenue}.

Between time steps 2 and 13, the \emph{Revenue} sub-agent decides to activate more servers. 
Yet, this decision is never taken by the aggregated RL agent until time step 13. 
The \textit{Running Costs} sub-agent, in contrast, regularly decides to turn off more servers in the second half of the example, when the request rate drops again. 
This is in direct contradiction to the decision of the \textit{Revenue} sub-agent. 

\begin{figure*}[ht]
    \centering
    		$
		\begin{array}{ccc}
    \includegraphics[width=.29\textwidth]{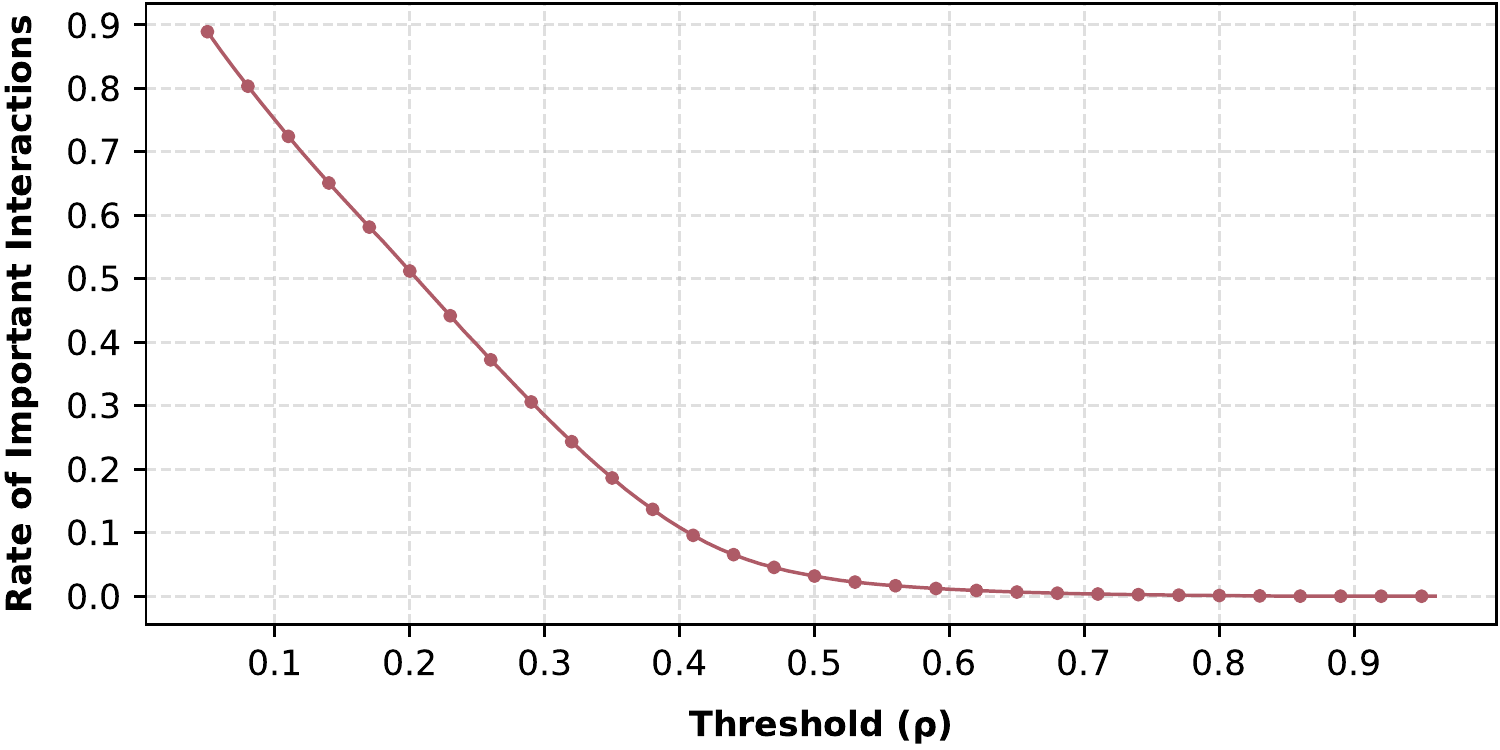} &	\includegraphics[width=.29\textwidth]{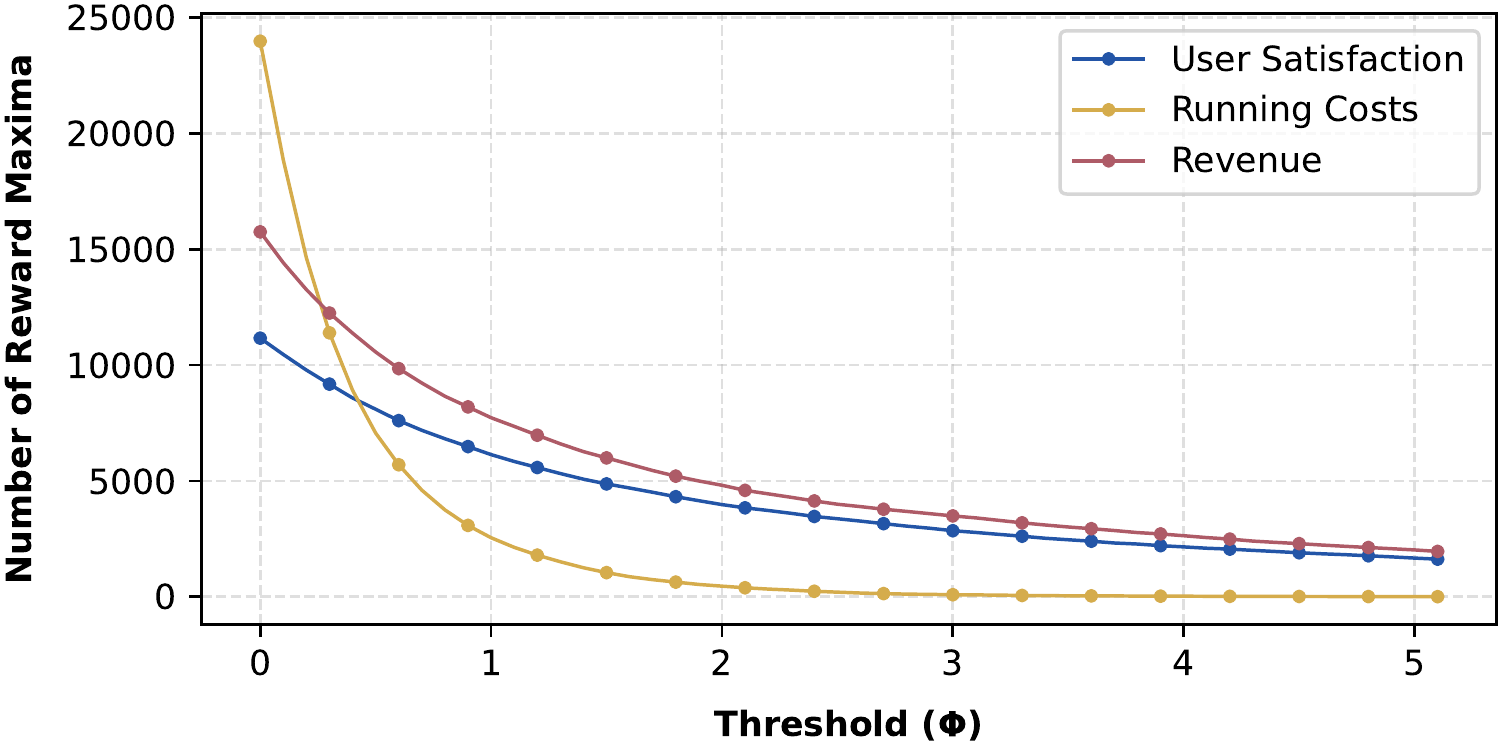} &		
	\includegraphics[width=.29\textwidth]{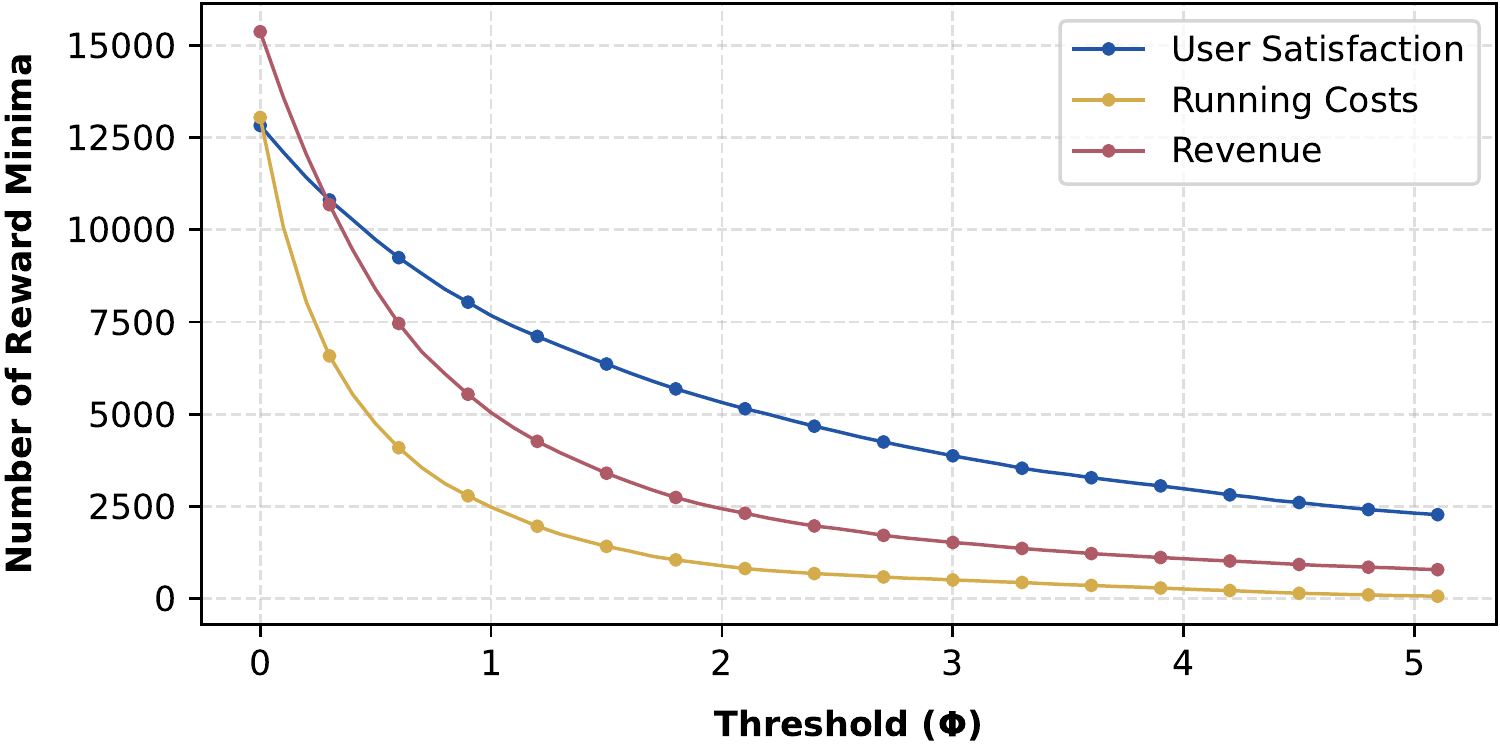}
\end{array}$
    \caption{Influence of threshold on number of DINEs shown in XRL-DINE dashboard}
%    \vspace{-1em}
    \label{fig:thresholds}
\end{figure*}

To better understand the aggregated agent's internal decision making when the aggregated decision changes to "Add Server", we look at the ``Reward Channel Dominance'' DINEs for time step 13, shown on the right hand side of Fig.~\ref{fig:example_3_top}. 
This action is the last action before actually adding another server, thus satisfying the repeated request of the sub-agent for \textit{Revenue}. 
In addition to the \textit{Revenue} sub-agent, the \textit{Running Costs} sub-agent also proposes an alternate action at this time step.
The ``Relative Reward Channel Dominance'' DINE shows that the sub-agent for \textit{User Satisfaction} has the greatest influence on the selected action \textit{No Adaptation}. 
The other two sub-agents each show visibly less reward channel dominance.  
This imbalance suggests a strongly biased action selection. 
It can also be seen that the chosen action is closely followed by the second-best action \emph{Add Server}. 
The alternative action \emph{Remove Server} proposed by the \textit{Running Costs} sub-agent is significantly worse, with less than half as much cumulative Relative Reward Channel Dominance, compared to the \emph{No Adaptation} action.

%\begin{figure}[thp]
%    \centering
%    \includegraphics[width=1\columnwidth]{fig/SWIM-2.pdf}
%    \vspace{-2em}
%    \caption{"Relative Reward Dominance" DINE for time step 13}
%    \label{fig:example_3_bottom}
%\end{figure}

%Regarding the "Reward Channel Extremum" DINE, we can observe: 
%At the beginning and at the end, the trace for the \textit{User Satisfaction} reward channel shows minima, which means that the sub-agent expects an increase in action-values in all possible subsequent states. 
%In contrast, the sub-agent for \textit{Revenue} expects a decrease in action-values in three of the first four steps. 
%The agent for \textit{Running Costs} determines a local minimum of action-values at the beginning and at the end of the sequence.

In the example, the aggregated agent decides to sacrifice \textit{Revenue} for higher \textit{User Satisfaction} and lower \textit{Running Costs}. Meanwhile, the sub-agent for \textit{Revenue} suggests alternative actions.
This is something one may expect when having some knowledge of the domain. 
These suggestions all relate to the \emph{Add Server} action. Adding more servers results in a lower average server utilization. 
This lower utilization would allow the dimmer to be raised again without compromising the dominant \textit{User Satisfaction} reward. 
Thus, from a domain perspective, these alternative actions make sense. 
The proposed alternative actions of the \textit{Running Costs} sub-agent also make sense from domain point of view, since removing servers intuitively leads to lower server costs.

Summarizing, DINEs in this example facilitate explicitly representing the agent's internal decision making trade-offs. 
From a domain point of view, one can think of three reasonable strategies for responding to the unanticipated increase in request rate shown in the example: (1) more servers can be added to ensure a high level of \textit{Revenue} and \textit{User Satisfaction} but at the expense of \textit{Running Costs}, (2) the dimmer value can be lowered to ensure a high level of \textit{User Satisfaction} and \textit{Running Costs} at the expense of \textit{Revenue}, or (3) no action at all can be taken at the expense of \textit{User Satisfaction}.

In the chosen example, the agent follows the second strategy. 
Developers of the self-adaptive system can evaluate this strategy and, if they are dissatisfied with the choice (and may consider it as buggy behaviour), they may change the reward function such as to prefer the alternative strategy suggested by the Revenue sub-agent.

To further support such debugging of the RL agent, ``Reward Channel Dominance'' DINEs can be leveraged to provide additional insights. 
For example, the developer can identify from the dashboard that \textit{User Satisfaction} alone is not sufficient for the agent to favor \emph{No Adaptation} over \emph{Add Server}. 
Instead, the relatively small influence of \textit{Running Costs} ultimately makes the difference. 
By investigating the ``Reward Channel Dominance'' DINEs of the selected actions, developers may determine that, from an agent's perspective, the first strategy listed above is rated second best by the agent.
This insight may then be used to adjust the structure or weighting of the reward function's components.

\subsection{Quantitative Results}
\label{sec:quantitative}

To complement the above results, we measure the number of DINEs shown to developers, thereby serving as indicator for the cognitive load.
We perform measurements for the different workload traces, covering a total of 62,000 timesteps.
In particular, we measure  how the number of DINEs depends on choosing the two thresholds: $\rho$ for ``Important Interaction'' DINEs and $\phi$ for ``Reward Channel Extremum'' DINEs.
To facilitate comparability of results, we use data from a single run of the RL agent, but filter accordingly based on the different thresholds.
Results are shown in Fig.~\ref{fig:thresholds}.
As can be seen, the thresholds allow tuning the rate of DINEs in a wide range; e.g., from close to zero up to 100\% in case of the $\rho$ threshold.
As the thresholds can be changed at runtime, developers may dynamically tune the rate of DINEs (1) based on their needs; e.g., coarse-grain observation vs. in-depth debugging, and (2) depending on the frequency of environment interactions.

\section{Discussion} 
\label{sec:discussion}

\subsection{Validity Risks} 
We used an actual self-adaptive system exemplar together with 30 GB of real-world workload traces when validating XRL-DINE.
Still results are only for a single system, which thus limits generalizability.

We tuned the hyperparameters of the RL algorithm experimentally, using educated guessing (e.g., comparable to~\cite{mnih_human-level_2015}). 
We purposefully did not perform extensive, exhaustive hyperparameter tuning, e.g., using grid search, because our aim was not to improve or compare the performance of existing RL approaches, but to validate how XRL-DINE may be used to generate explanations for RL decisions.

We measured the number of explanations (DINEs) generated depending on the parameters of XRL-DINE.
Such measurement only served as a surrogate (i.e., indirect indicator) for measuring cognitive load.
The results presented thus should be complemented by future user studies, which allow assessing the actual utility of the information given to humans by XRL-DINE.

\subsection{Current Limitations} 
\runin{Understandability of explanations}  is a general concern for any explainable ML technique.
XRL-DINE may generate difficult to understand explanations in two situations:
First, the reward function may have been decomposed incorrectly or at least non-optimally. 
This may occur, for example, if the reward channels have direct dependencies or cannot be clearly separated. 
It then becomes mentally difficult to grasp the meaning of a reward channel, because rewards may be influenced by several factors. 
Second, environment dynamics may delay the effects of adaptations. 
As an example, in SWIM there is a delay before a new server has booted after executing the \emph{Add Server} action.
In contrast, lowering the dimmer value has an immediate effect on latency and thus on \textit{User Satisfaction} rewards and might therefore be favored. 
Such timing-related differences may make interpreting explanations more difficult.

\runin{Policy-based RL} is another class of modern deep RL algorithms used for self-adaptive systems~\cite{CAISE2020}.
XRL-DINE is not applicable to policy-based RL, because the underlying XRL techniques require access to the value-function $Q(S,A)$, which does not exist in policy-based RL.

\runin{Explanations for collaborative adaptive systems} (e.g., see~\cite{DAngeloGGGNPT19}) requires the extension of XRL-DINE. 
While XRL-DINE can effectively explain the decisions of a single RL agent, XRL-DINE does not consider the decisions of other RL agents during explanations. 
Similarly to the aforementioned situations in which XRL-DINE may generate difficult to understand explanations, the same may happen if XRL-DINE is directly applied in the collaborative setting.

\runin{Addressing the findings of multidisciplinary research} requires generating explanations that are (1)~contrastive, (2)~selective, (3)~causal, and (4)~social~\cite{miller_explanation_2019}.
While XRL-DINE addresses the first two points, the remaining two ones are not yet addressed, i.e., XRL-DINE is not yet causal~\cite{Dusparic_2022} and not yet conversational~\cite{PolDD19}.

\section{Related Work}
\label{sec:RelatedWork}

While generic explainable RL approaches are discussed in recent overview papers (such as~\cite{heuillet_explainability_2020, puiutta_explainable_2020}), these do not specifically address self-adaptive systems. 
We thus focus the following discussion on solutions that specifically provide explanations for self-adaptive systems.
Here, the following four main groups of work can be identified.

The first group of work proposes using a temporal graph model as a central artifact to derive explanations~\cite{garcia-dominguez_towards_2019, UllauriGBZZBOY22}. 
On the one hand, such a model may be used for so called forensic self-explanation.
This means that the model may be queried via a dedicated query language. 
On the other hand, such a model may be used for so called live self-explanation.
Here, one can submit queries to the running system and be presented with a live visualization. 
The underlying temporal model is kept up to date at runtime (i.e., employed as a model@runtime).
The approach is comparable to the Interestingness Elements described in Sect.~\ref{sec:Foundations} since explanations are generated based on execution traces. 
However, in contrast to XRL-DINE, interesting interactions must be extracted by manually writing queries using the provided query language. 
As follow-up work, suggestions for automating the selection of interesting interaction moments are proposed.
In~\cite{UllauriGBZZBOY22} this is fully automated by using complex-event-processing.
While the aim of selecting interesting interactions in~\cite{UllauriGBZZBOY22} is to keep the size of the models@runtime manageable, the aim of XRL-DINE is to reduce the cognitive load of developers. 
Compared to previous work in this group,~\cite{UllauriGBZZBOY22} stands out in providing explanations for RL decisions.
While the paper hints at the possibility of using model@runtime queries to realize reward decomposition, it differs from XRL-DINE in that the combination of Interestingness Elements and Reward Decomposition is not considered.

The second group of work uses goal-based models at runtime~\cite{bencomo_self-explanation_2012, welsh_self-explanation_2014}. 
In~\cite{bencomo_self-explanation_2012}, higher-level system traces are used as explanations.
%Again, this can be considered similar to the idea of Interestingness Elements (see Sect.~\ref{sec:Foundations}).
In~\cite{welsh_self-explanation_2014} a domain-specific language for providing explanations in terms of the satisficement of softgoals is introduced. 
%In this regard, explanations are comparable to the generic Reward Decomposition explanations described in Sect.~\ref{sec:Foundations} in that the explanations refer to competing goal dimensions. 
Other than XRL-DINE, these techniques require making assumptions about the environment dynamics at design time, which can be a source for error due to design time uncertainty~\cite{weyns_perpetual_2013}.
Also, in contrast to XRL-DINE, this group of work does not explicitly consider RL.

The third group of work uses interaction data collected at runtime to generate explanations in the form of provenance graphs~\cite{reynolds_automated_2020}. 
A provenance graph contains information and relationships that contributed to the existence of a piece of data. 
By keeping a history of different versions of the provenance graph, it is possible to determine at runtime \textit{if} and \textit{how} the model has changed (using model versioning) and \textit{who} has changed the model and \textit{why} (using the provenance graph).
Provenance graphs quickly can become too complex to be meaningfully interpreted by humans, thus a dedicated query language was introduced that allows extracting information of interest.
Again, in contrast to XRL-DINE, this group of work does not consider RL.

The fourth group of work takes a fundamentally different view on explainability~\cite{li_hey_2021}. 
A formal framework is proposed in which explainability is not provided externally but is considered a concrete tactic of the self-adaptive system. 
In uncertain or difficult situations, the self-adaptive system can ask assistance from a human operator in making a decision rather than acting itself. 
%Specifically, the overall system is modeled as a turn-based, stochastic multiplayer game in which three players participate. 
%These players are (1) the actual self-adaptive system, (2) the environment, and (3) the operator. 
%This game is then analyzed using a probabilistic model checker to determine when the involvement of a human operator is necessary.
To prevent the operator from being permanently consulted, there is a cost to using this tactic that must be accounted for.
% by the model checker. 
In this respect, this approach is similar to XRL-DINE, which aims to reduce the cognitive burden of the human. 
However, the motivation in XLR-DINE is the limited cognitive ability of the human, while in~\cite{li_hey_2021} it is the time delay caused by involving an operator.
Again, RL is not explicitly considered.

\section{Conclusion and Perspectives}
\label{sec:ConclusionAndPerspectives}

We introduced XRL-DINE, a technique for explaining Deep RL systems for self-adaptive systems.
XRL-DINE enhances and combines two existing explainable RL techniques from the machine learning literature, overcoming the respective limitations of the individual techniques.  
We described the prototypical implementation of XRL-DINE using a state-of-the-art deep RL algorithm, serving as proof-of-concept.
Qualitative results demonstrate the usefulness of XRL-DINE, while quantitative results show the potential reduction in cognitive load required to interpret explanations.
%In its current state, XRL-DINE can be used by developers to gain insight and spot errors in the decision-making process of RL-based self-adaptive systems.

As future work, we plan enhancing XRL-DINE by addressing its current limitations.
In addition, it may be interesting to provide actionable suggestions to developers on how to change the reward functions to obtain different adaptation decisions.
Finally, we plan performing user studies to evaluate XRL-DINE's usefulness in practice.
% (e.g., following~\cite{Doshi-VelezK17}).

\runin{Acknowledgments.} We cordially thank the anonymous reviewers for providing constructive comments for the final version of the paper. 
Research leading to these results received funding from the EU’s H2020 R\&I programme under grants  780351 (ENACT) and 	871493 (DataPorts).

%\vspace{1em}
%{\emph{Acknowledgment:} 
%We thank the anonymous reviewers for their constructive comments.
%The research leading to these results received funding from the European Union's Seventh Framework Programme FP7/2007-2013 under grant agreement 610802 (CloudWave), from the European Union’s Horizon 2020 research and innovation programme under grant agreement no. 731932 (TransformingTransport), as well as from the EFRE co-financed operational program NRW.Ziel2 under grant agreement 005-1010-0012 (LoFIP).}

\bibliographystyle{IEEEtran}
\bibliography{IEEEabrv,acsos2022,acsos2022extra}

\end{document}